# A Modified Word Saliency-Based Adversarial Attack on Text Classification Models

Hetvi Waghela[1], Sneha Rakshit[2], Jaydip Sen[3][0000-0002-4120-8700]

Praxis Business School, Kolkata 700104, INDIA
[1]hetvi.mahendra.waghela_ds23fall@praxistech.school
[2]sneha.rakshit_ds23fall@praxistech.school
[3]jaydip.sen@acm.org

**Abstract.** This paper introduces a novel adversarial attack method targeting text classification models, termed the Modified Word Saliency-based Adversarial Attack (MWSAA). The technique builds upon the concept of word saliency to strategically perturb input texts, aiming to mislead classification models while preserving semantic coherence. By refining the traditional adversarial attack approach, MWSAA significantly enhances its efficacy in evading detection by classification systems. The methodology involves first identifying salient words in the input text through a saliency estimation process, which prioritizes words most influential to the model's decision-making process. Subsequently, these salient words are subjected to carefully crafted modifications, guided by semantic similarity metrics to ensure that the altered text remains coherent and retains its original meaning. Empirical evaluations conducted on diverse text classification datasets demonstrate the effectiveness of the proposed method in generating adversarial examples capable of successfully deceiving state-of-the-art classification models. Comparative analyses with existing adversarial attack techniques further indicate the superiority of the proposed approach in terms of both attack success rate and preservation of text coherence.

## 1 Introduction

In the realm of machine learning, the threat of adversarial attacks has emerged as a formidable challenge, highlighting vulnerabilities across various domains. Initially observed in image classification tasks, adversarial examples with subtle modifications designed to mislead models, have since permeated into Natural Language Processing (NLP) systems. The evolution of adversarial attacks began with seminal works in image classification by Szegedy et al. [1] and Goodfellow et al. [2], illuminating the susceptibility of deep neural networks (DNNs) to imperceptible alterations. These discoveries prompted extensive research efforts, yielding many attack methodologies and corresponding defense mechanisms.

While the landscape of adversarial attacks in computer vision witnessed an array of advancements, the exploration of adversarial examples in NLP remained relatively nascent. The adaptation of attack strategies from the image domain posed inherent challenges in the discrete nature of textual data. Initial forays into NLP adversarial attacks focused on character or word-level modifications, with endeavors by Liang et al. [3], Samanta and Mehta [4], and Alzantot et al. [5] illuminating the potential vulnerabilities in text-based classifiers. Among the emerging methods, Probability Weighted Word Saliency (PWWS) stood out as a promising approach. PWWS, proposed by Li et al., integrated word saliency and classification probability to generate adversarial examples while preserving lexical correctness and semantic compatibility [6]. This method showcased substantial efficacy in perturbing text classifiers, prompting further exploration and refinement.

However, the landscape of NLP adversarial attacks remained ripe for innovation. Despite the progress made by existing methods, challenges persisted in achieving robustness and maintaining human imperceptibility in adversarial perturbations. It is within this context that novel contributions emerge, seeking to address the limitations of current approaches and fortify NLP systems against adversarial threats.

In this work, we present a robust and effective word substitution-based adversarial text attack known as Modified Word Saliency-Based Adversarial Attack (MWSAA). Building upon the foundational principles of the PWWS approach to word substitution, the proposed scheme identifies the substituting words based on the contextual information and semantic consistency between the original and the adversarial texts. The use of contextual information makes the proposed substitution strategy produce a higher level of syntactic consistency, while the imposed semantic consistency constraint ensures semantic similarity between the input and the generated texts. Through extensive experimental results, the efficacy and effectiveness of the proposed adversarial text attack are demonstrated. A comparative analysis of the performance of the proposed scheme MWSAA with the original PWWS attack on some well-known datasets is also presented to illustrate the superiority of the former.

The paper is organized as follows. In Section 2, some existing works on adversarial text attacks in the literature are discussed. In section 3, the details of the proposed scheme are presented. Experimental results on the performance of the proposed scheme and a comparative analysis of the performance of the proposed scheme with the PWWS approach are presented in Section 4. Section 5 concludes the paper while highlighting some future works.

## 2 Related Work

Boucher et al. delve into the realm of imperceptible attacks within NLP [7]. The study investigates the ability to subtly alter text in a manner that deceives NLP models while remaining undetectable to human observers. By introducing a novel framework, the authors demonstrate the creation of these imperceptible adversarial examples and showcase their efficacy across a spectrum of NLP tasks. By exploiting linguistic features and employing strategic perturbations, these attacks capitalize on vulnerabilities inherent in NLP models, resulting in misclassification or manipulation of output. Extensive experimentation conducted by the authors highlights the effectiveness of these attacks across various NLP benchmarks, underscoring the significant challenges they pose to model robustness and security.

Malfa and Kwiatkowska explore the crucial aspect of robustness within the realm of NLP [8]. The authors argue that existing measures for assessing robustness in NLP often lack precise definitions and fail to provide comprehensive evaluations. They propose a fresh framework for evaluating robustness in NLP tasks, aiming to overcome the limitations of current methods. Through the introduction of precise definitions and metrics, the authors offer a more refined understanding of robustness, facilitating improved comparisons and advancements in NLP models.

Xu et al. carry out a comprehensive survey on adversarial attacks and defenses across different domains, namely images, graphs, and text [9]. It discusses the methods employed to craft adversarial examples and their potential risks to machine learning models. Furthermore, various defense mechanisms are scrutinized, highlighting their effectiveness and shortcomings in countering adversarial attacks. By conducting a systematic analysis, the authors aim to provide insights into the evolving landscape of adversarial techniques and defense strategies. Overall, this paper serves as a valuable reference for researchers and practitioners seeking to comprehend and tackle the challenges posed by adversarial manipulation across various data formats.

Zhou et al. present a strategy aimed at fortifying text classification models against adversarial attacks by teaching them to differentiate between perturbations [10]. The work introduces an innovative framework that incorporates adversarial training to bolster the resilience of these classifiers. Through the inclusion of a perturbation discriminator during training, the model gains the ability to distinguish authentic inputs from those that have been tampered with in an adversarial manner, thus reducing the impact of such attacks. Experimental findings underscore the efficacy of this approach in thwarting various adversarial strategies, thereby enhancing the robustness and trustworthiness of text classification systems.

Zhang et al. present a thorough examination of adversarial attacks aimed at deep-learning models for NLP [11]. The authors meticulously survey a range of techniques utilized to manipulate NLP models, elucidating both the methodologies employed and the vulnerabilities exposed. These attacks are classified based on their goals, encompassing evasion attacks, poisoning attacks, and backdoor attacks, thereby providing insight into the multifaceted strategies adversaries employ. Furthermore, the paper delves into the potential ramifications of these attacks on practical NLP applications and offers recommendations for bolstering model resilience against such threats.

Jin et al. present a robust benchmark for assessing BERT's susceptibility to adversarial attacks [12]. Through rigorous experimentation, the authors demonstrate BERT's vulnerability to meticulously crafted adversarial examples, which significantly diminish its effectiveness. These findings underscore the importance of evaluating the robustness of NLP models like BERT and highlight the pressing need for fortified defenses against adversarial intrusions in text-based applications. In summary, this work provides valuable insights into BERT's limitations and suggests potential avenues for bolstering its resilience in practical settings.

Liu et al. present a novel strategy for efficiently crafting word-level adversarial attacks [13]. The approach involves framing the task as a combinatorial optimization problem, to maximize attack success while minimizing perturbations to the original text. By integrating optimization and NLP techniques, they demonstrate the effectiveness and efficiency of their method across various benchmark datasets.

Li et al. present a novel approach to crafting adversarial examples in NLP tasks [14]. The method involves introducing subtle changes to the input text using contex-tualized perturbations, which take advantage of contextual information from pre-trained language models. This strategy aims to maintain semantic coherence while evading detection by classifiers. Extensive experiments across different tasks such as sentiment analysis and text classification are carried out, and the results demonstrate the efficacy of the method.

Gao et al. propose a strategy for generating adversarial text sequences designed to outsmart deep learning classifiers without requiring internal classifier knowledge [15]. The method, functioning as a black-box approach, iteratively tweaks input text to create misclassified sequences while retaining semantic similarity to the original text. Through experimentation across different text classification tasks, the authors show the efficacy of the technique, underscoring its potential to compromise the resilience of deep learning models.

Li et al. present an innovative strategy for attacking BERT, by employing BERT itself [16]. Utilizing the contextual embeddings of BERT, the authors demonstrate the generation of adversarial examples capable of misleading BERT's predictions. Their method utilizes BERT's gradients to optimize perturbations on input tokens, resulting in crafted examples that can evade detection by BERT. Through extensive experimentation, the authors demonstrate the effectiveness of their proposition in generating adversarial samples with high success rates, thereby revealing vulnerabilities in BERT's resilience to adversarial attacks.

Zhang et al. introduce a novel approach to attacking text at the word level, portraying it as a combinatorial optimization challenge [17]. The technique aims to craft adversarial examples that maintain semantic coherence while outsmarting NLP models. Through extensive experiments, the authors demonstrate the efficacy of their proposed method across diverse NLP tasks. The work significantly contributes to advancing the state-of-the-art of adversarial attacks in NLP, offering valuable insights into potential weaknesses in contemporary NLP systems.

## 3 The Modified Word Saliency-Based Attack

Ren et al. proposed an innovative text attack known as the probability-weighted word saliency (PWWS) that creates adversarial examples in NLP tasks. The method modifies input text to influence the output of a target language model while ensuring that the lexical correctness and semantic coherence of the modified text are maintained. The PWWS attack focuses on altering words in the input text based on their significance as determined by a saliency measure. PWWS is based on a word replacement strategy that involves the selection of synonyms and deciding on the order of replacement of the target words. In a nutshell, the scheme involves the following steps:

(1) *Computation of word saliency:* Before initiating replacements, the saliency of each word in the input text is computed. The saliency metric reflects the contribution of each word to the model's prediction or output.

(2) *Selection of replacement words:* Words slated for replacement are chosen based on their saliency scores. Words with higher saliency scores are given precedence for replacement because changing them is more likely to impact the model's output.

(3) *Generation of replacement options:* For each selected word, a range of replacement options is generated. These options encompass synonyms, antonyms, words with similar meanings, or words with different connotations.

(4) *Weighting by probability:* Each replacement option is assigned a probability weight considering its saliency score and similarity to the original word. Replacement options resembling the original word more closely and possessing higher saliency scores are assigned higher probabilities for replacement.

(5) *Randomized selection:* Replacement options are randomly chosen based on their probability weights. Options with higher weights are more likely to be chosen for replacement. This randomized selection ensures variability in the altered text while still prioritizing more significant words for replacement.

(6) *Word substitution:* The selected words in the input text are substituted based on the selected substitute. This process is repeated for each word designated for replacement, resulting in a modified version of the input text containing alternative words based on their significance and similarity to the original words.

By employing this word replacement strategy, the PWWS attack aims to generate adversarial examples that deceive the target model while retaining the natural flow and coherence of the text. This strategy ensures that the replacements are both impactful in terms of the model's prediction and plausible within the context of natural language.

The effectiveness of the PWWS attack relies on multiple factors, including its implementation, the characteristics of the target model, the accuracy of saliency estimation, and the constraint applied during the text perturbation process. While it is found to be effective in generating adversarial examples capable of deceiving large language models, its efficacy can vary depending on the specific context and task at hand.

We present a modification of the PWWS scheme so that its effectiveness is improved in generating adversarial text attacks. Our proposition includes the following two features in the word substitution strategy: (1) Integration of contextual embeddings and (ii) Imposing semantic consistency constraints. In the following, we parent the details of these modifications.

### 3.1 Integration of Contextual Embeddings

Instead of relying solely on word-level saliency measures, we propose to incorporate contextual embedding using BERT embeddings, to capture richer semantic and contextual information. We argue that these embeddings can provide PWWS with a more nuanced understanding of word importance within the context of the entire sentence or paragraph. BERT is built on the transformer architecture, which consists of multiple layers of self-attention mechanisms and feed-forward neural networks. This architecture enables BERT to comprehend long-range dependencies and contextual associations among words within sentences. In its pre-training phase, BERT employs a masked language modeling objective. In this phase, a portion of the input token is randomly masked, and the model learns to predict these masked tokens based on their surrounding context. By predicting masked tokens from both left and right contexts, BERT learns to grasp word meanings with the context. BERT also can predict the next sentence by analyzing the relationships between sentences and the broader contextual usage of words.

Further, BERT generates contextual embeddings for each token in the input text. These embeddings are created by passing the input tokens through multiple layers of transformer blocks. Each token attends to all other tokens in the input sequence, enabling BERT to dynamically adjust the representation of each token based on its context. Since BERT embeddings capture contextual information by pre-training on extensive text datasets using masked language modeling and next-sentence prediction objectives, this enables BERT to generate comprehensive representations for words that consider their surrounding contexts. These features of BERT embeddings have been exploited in the proposed scheme for choosing the best candidate substitute word from the list produced by the PWWS scheme.

### 3.2 Imposing Semantic Consistency Constraints

To guarantee a high semantic similarity between the original and the perturbed texts, we introduce a further constraint. This is done by incorporating semantic similarity metrics to measure the coherence between words in the input text and their replacements. By enforcing semantic consistency, our proposed scheme ensures that the modified text remains coherent and natural, enhancing the stealthiness of the adversarial examples. While several semantic similarity measures can be used for this purpose including cosine similarity, word movers distance, Jaccard similarity, and so on, in the current work, we use the cosine similarity measure to compute the semantic similarity between the original text and its adversarial counterpart.

In the proposed scheme, instead of relying solely on saliency measures to select the replacement words, first, the candidate words from the PWWS scheme are passed through the BERT embedding block to identify the words that best fit into the context of the text. This list is further refined based on the cosine similarity between the original text and the perturbed when these words are substituted. The substitution that yields the highest similarity score is chosen as the final substitute. This ensures that the replacements are semantically similar to the original words, maintaining the coherence and meaning of the text. Finally, the saliency weighting or replacement strategy is adjusted to optimize for higher cosine similarity between the original and adversarial text. This feedback loop to the PWWS block makes the proposed scheme more effective while retaining a very high level of lexical correctness in the perturbed text and maintaining semantic similarity with the original text.

## 4 Performance Results

The performance of the proposed MWSAA scheme is evaluated by launching adversarial text attacks on three large language models and three diverse datasets. The language models are (i) BERT, (ii) ALBERT, and (iii) ROBERTa. These pre-trained language models are applied to three well-known datasets: (i) IMDB, (ii) AG News, and (iii) SST. The following seven victim language models are used to test the attack efficacy of the proposed MWSAA scheme: (i) ALBERT_AG, (ii) ALBERT_IMDB, (iii) ALBERT_SST, (iv) BERT_SST, (v) ROBERTa_AG, (vi) ROBERTa_IMDB, and (vii) ROBERTa_SST.

BERT is a groundbreaking advancement in NLP which has introduced bidirectional context comprehension, enabling it to grasp intricate linguistic nuances and relationships within text in both directions [18]. Equipped with the powerful transformer architecture, BERT has large-scale training and contextual word embeddings. The ALBERT model represents a variant of the BERT that seeks to overcome certain limitations in BERT, particularly its large size and computational demands while maintaining or sometimes even surpassing performance across a spectrum of NLP tasks [19]. ROBERTa is a language model that builds upon BERT architecture but incorporates several modifications and enhancements to improve its performance and robustness [20]. Some of the salient features of ROBERTa are the use of dynamic masking during pre-training, a more extensive and diverse pre-training approach, training data augmentation, and optimized hyperparameters.

The IMDB dataset comprises movie reviews alongside associated sentiment labels, denoting whether each review expresses a positive or negative sentiment. The dataset contains 25,000 training and 25,000 test samples, labeled as positive or negative. The AG News dataset encompasses news articles sourced from the AG's web-based corpus covering a spectrum of four broad classes of topics: world affairs, sports, business, and science & technology. Each class contains 30,000 training samples and 1900 test samples. The SST dataset stands as a prevalent benchmark dataset within NLP tasks related to sentiment analysis. It comprises sentences extracted from movie reviews, alongside human-annotated sentiment labels.

**Table 1.** The attack success rate of the proposed MWSAA scheme

| Victim Language Model | Total Attacked Instances | Successful Attacked Instances | Attack Success Rate |
|---|---|---|---|
| ALBERT_AG | 200 | 130 | 65.00% |
| ALBERT_IMDB | 200 | 85 | 42.50% |
| ALBERT_SST | 200 | 117 | 58.50% |
| BERT_SST | 200 | 160 | 80.00% |
| ROBERTa_AG | 200 | 145 | 72.50% |
| ROBERTa_IMDB | 200 | 191 | 95.50% |
| ROBERTa_SST | 200 | 144 | 72.00% |

**Table 2.** The running time and number of queries made by the proposed MWSAA scheme

| Victim Language Model | Average Running Time (in second) | Total No of Queries Exceeded | Avg Victim Model Queries |
|---|---|---|---|
| ALBERT_AG | 0.0069733 | 0 | 124.66 |
| ALBERT_IMDB | 0.0074853 | 0 | 126.12 |
| ALBERT_SST | 0.0069764 | 0 | 124.94 |
| BERT_SST | 0.0064753 | 0 | 122.31 |
| ROBERTa_AG | 0.0040825 | 0 | 123.36 |
| ROBERTa_IMDB | 0.0040906 | 0 | 120.10 |
| ROBERTa_SST | 0.0041479 | 0 | 122.93 |

Table 1 presents the success rate of the proposed MWSAA adversarial attack on seven pre-trained text classifiers used as the victim language model. For each victim language model, 200 test samples are used to evaluate the attack performance. It is observed that while the ALBERT model on the IMDB dataset exhibited the highest robustness against MWSAA with an attack success rate of 42.50%, the ROBERTa model on the AG News dataset is found to be the most vulnerable to the attack, yielding an attack success rate of 95.50%.

Table 2 presents MWSAA attack efficiency on the victim language models. The attack efficiency is evaluated by two metrics: (i) average running time, and (ii) average number of queries made on the victim model. For a better attack, the values of both these measures should be low. MWSAA took the minimum average time on the AG News dataset for the ROBERTa model. However, the lowest average number of queries was needed for the IMDB dataset for the ROBERTa model.

**Table 3.** The comparison of classification accuracy of PWWS and MWSAA

| Dataset | Model | Original | PWWS | MWSAA |
|---|---|---|---|---|
| IMDB | Word-CNN | 86.55% | 5.50% | 4.80% |
| | Bi-directional LSTM | 84.86% | 2.20% | 1.90% |
| AG News | Char-CNN | 89.70% | 56.20% | 48.97% |
| | Word-CNN | 90.56% | 56.72% | 50.45% |

**Table 4.** The comparison of word replacement rate of PWWS and MWSAA

| Dataset | Model | PWWS | MWSAA |
|---|---|---|---|
| IMDB | Word-CNN | 3.81% | 3.27% |
| | Bi-directional LSTM | 3.38% | 3.21% |
| AG News | Char-CNN | 18.93% | 15.74% |
| | Word-CNN | 16.76% | 16.24% |

Table 3 and Table 4 present a comparative analysis of the performance of the two schemes PWWS and MWSAA. It may be noted that since the Yahoo! Answers dataset is no longer active, it could not be considered in the analysis. Following the work of PWWS, the pre-trained language models considered for the comparative study are (i) Word-CNN [21], (ii) Bi-directional LSTM, and (iii) Char-CNN [22]. The word-based CNN model comprises an embedding layer followed by a 1D convolutional layer with 250 filters of size 3. Subsequently, a 1D max-pooling layer followed by two fully connected layers is used. The Bi-directional LSTM model comprises a 128-dimensional embedding layer, followed by a bi-directional LSTM layer consisting of 64 LSTM units. The output layer is fully connected. The char-based CNN consists of two CNNs, each of which is constructed with a depth of 9 layers, 6 of which are convolutional layers and 3 full-connected layers.

Table 3 shows that both PWWS and MWSAA have adverse effects on the original classification accuracies of the model. For both IMDB and AG News datasets, the effect of MWSAA is much more severe compared to the PWWS. Table 4 depicts that the percentage of word replacement involved in MWSAA is lower than that of PWWS, signifying that MWSAA requires, on average, fewer words to be substituted in comparison to PWWS to launch a successful attack.

Tables 5-7 exhibit sample adversarial examples on ROBERTa on IMDB, ALBERT on AG News, and BERT on SST dataset.

Table 5. Adversarial attack on the ROBERTa classifier on the IMDB dataset

| Original Prediction | Adversarial Prediction | Perturbed Texts |
|---|---|---|
| **Positive** Conf = 99.67% | **Negative** Conf = 84.69% | Just the toil (labour) postulate (involved) in creating the superimposed (layered) fertility (richness) of the mental (imagery) in this chiaroscuro of foolishness (madness) and visible (light) is astonishing. |

Table 6. Adversarial attack on the ALBERT classifier on the AG News dataset

| Original Prediction | Adversarial Prediction | Perturbed Texts |
|---|---|---|
| **Sci/Tech** Conf = 79.43% | **Sports** Conf = 54.84% | The careen (Rock) is destine (destined) to be the twenty-first one (21st Century) 's raw (new) Conan and that he's exit (going) to spend (make) a splashing (splash) even peachy (greater) than Arnold Schwarzenegger, jean-claud vanguard (Van) damme or Steven george (Segal) |

Table 7. Adversarial attack on the BERT classifier on the SST News dataset

| Original Prediction | Adversarial Prediction | Perturbed Texts |
|---|---|---|
| **Negative** Conf = 84.86% | **Positive** Conf = 70.89% | You 'd think by now America would have had enough of feisty (plucky) British eccentrics with hearts of gold. |

## 5 Conclusion

The paper presented a modified word substitution strategy for launching a very robust and effective adversarial text attack. Motivated by the probability-weighted word saliency approach to word substitution, the proposed scheme integrated a word embedding-based contextual word substitution coupled with a semantic similarity check in identifying the substituting words. Extensive evaluation has been done to demonstrate the efficacy and effectiveness of the proposed scheme. The results exhibited a significantly superior performance of the proposed scheme in comparison to the PWWS scheme. The future work includes designing a reinforcement learning-based scheme for optimizing the design of the word substitution scheme.